# Spiking SiamFC++: Deep Spiking Neural Network for Object Tracking

Shuiying Xiang, Tao Zhang, Shuqing Jiang, Yanan Han, Yahui Zhang, Chenyang Du, Xingxing Guo,
Licun Yu, Yuechun Shi and Yue Hao

*Abstract*— **Spiking neural network (SNN) is a biologically-plausible model and exhibits advantages of high computational capability and low power consumption. While the training of deep SNN is still an open problem, which limits the real-world applications of deep SNN. Here we propose a deep SNN architecture named Spiking SiamFC++ for object tracking with end-to-end direct training. Specifically, the AlexNet network is extended in the time domain to extract the feature, and the surrogate gradient function is adopted to realize direct supervised training of the deep SNN. To examine the performance of the Spiking SiamFC++, several tracking benchmarks including OTB2013, OTB2015, VOT2015, VOT2016, and UAV123 are considered. It is found that, the precision loss is small compared with the original SiamFC++. Compared with the existing SNN-based target tracker, e.g., the SiamSNN, the precision (succession) of the proposed Spiking SiamFC++ reaches 85.24% (64.37%), which is much higher than that of 52.78% (44.32%) achieved by the SiamSNN. To our best knowledge, the performance of the Spiking SiamFC++ outperforms the existing state-of-the-art approaches in SNN-based object tracking, which provides a novel path for SNN application in the field of target tracking. This work may further promote the development of SNN algorithms and neuromorphic chips.**

*Index Terms*— **Spiking neural network, Siamese network, object tracking, supervised learning, surrogate gradient.**

## I. Introduction

THE artificial neural networks (ANN) have made tremendous progress and have achieved impressive performance in various artificial intelligence (AI) tasks such as image classification and localization, speech recognition, data centers, and autonomous driving. However, as the memory and central processing units of the conventional von Neumann computing platforms are physically-separated, operating these ANNs with conventional computers pose significant energy demands due to the data transfer. Thus, it is urgent to develop novel computing platforms to meet the requirement of energy efficiency. Neuromorphic computing platform, which is inspired by the brain computing, has become one of the leading candidates to overcome the von-Neumann bottleneck [1]. Compared with the continuous-value based ANNs, the spiking neural network (SNN) models are more biologically-plausible and exhibit event-driven nature and low power consumption [1-3]. Running SNNs on neuromorphic hardware with reduced energy requirements are appealing for the resource- and energy-constrained applications [4-11]. Note that, in a SNN, the data is represented and propagated by discrete spikes. Thus, the training of SNN is difficult as the spike events are non-differentiable. Yet now, SNNs are still behind their ANN counterparts in terms of accuracy for most learning tasks.

Tremendous efforts have been devoted towards designing unsupervised and supervised algorithms for training SNNs. For instance, spike-timing-dependent plasticity (STDP) is a representative unsupervised learning rule [12-16]. There are also many supervised learning algorithms to train shallow SNNs [17-20]. Besides, to solve complex AI tasks, supervised training methods for multilayer or deep SNN have also been reported [21-24]. For instance, the "ANN-to-SNN conversion" method is a popular training approach, which converts a well-trained ANNs to SNNs instead of direct training [25-27]. Such conversion method exhibits energy efficiency and high performance that inherited from both the SNNs and ANNs [25-27]. Note that, in such a method, the firing rate of a spiking neuron, e.g., an Integrate-and-Fire (IF) neuron, is adopted to represent information in SNN, which is similar to the output of Rectified Linear Unit (ReLU) function in ANN. Namely, the rate encoding rather than the temporal encoding is adopted, which requires lots of time steps and results in high latency. In addition, the conversion method is difficult to be applied to deep SNN because the conversion error will accumulate layer by layer. To reduce the SNN latency, the surrogate gradient learning methods have been widely adopted in deep SNN in recent years [28-31]. For this method, the derivation of an approximate differentiable function was calculated to replace the derivation of the spiking event. Different from the conversion method, this surrogate gradient learning strategy with direct training requires less inference time-steps and thus

Manuscript received Sep 6, 2022. This work is supported by the National Key Research and Development Program of China (2021YFB2801901, 2021YFB2801902, 2021YFB2801904).

S. Xiang is with State Key Laboratory of Integrated Service Networks, Xidian University, Xi'an 710071, China; and also with State Key Discipline Laboratory of Wide Bandgap Semiconductor Technology, Xidian University, Xi'an 710071, China (email: syxiang@xidian.edu.cn).

T. Zhang, S. Jiang, C. Du, Y. Han Y. Zhang, X. Guo, are with State Key Laboratory of Integrated Service Networks, Xidian University, Xi'an 710071, China.

L. Yu is with CCCC First Highway Consultants Co., Ltd, Xi'an 710075, China; and also with School of Highway, Chang'an University, Xi'an 710064, China.

Y. Shi is with Yongjiang laboratory, No. 1792 Cihai South Road Ningbo, 315202, China. (e-mail: yuechun-shi@ylab.ac.cn).

Y. Hao is with State Key Discipline Laboratory of Wide Bandgap Semiconductor Technology, Xidian University, Xi'an 710071, China. (e-mail: yhao@xidian.edu.cn).

is suitable for energy-efficient applications of deep SNN. With these impressive supervised training approaches, the SNN models have been successfully used in various applications including image classification, object recognition and detection [15, 23, 25, 31-33].

Recently, object tracking based on deep learning have emerged and are widely used in security monitoring, unmanned driving, automatic robot navigation, smart home and other fields due to their outstanding tracking performance [34-35]. For the object tracking task, it actually includes both the classification and localization tasks. One of the main challenges in object tracking is to design an object tracker that is robust to appearance changes including occlusions, motion blur, and background clutter [35]. Recently, there are many reported object trackers based on deep learning that achieve impressive tracking performance. For example, fully-Convolutional Siamese Networks (SiamFC), which extracts feature with a convolutional neural network (CNN), performs convolution operation between two feature maps, and determines the location of the object through the response score map [36]. Subsequently, SiamRPN [37], SiamRPN++ [38], SiamFC++ [39] were further developed with improved performance in terms of speed, accuracy, and robustness. However, these methods require large amounts of computation and high resource consumption, which hinders their applications in AI chips and other devices with limited resources.

Very recently, SNN models have shown great potential as a solution in the object tracking field [40-43]. For instance, the SiamSNN, which represented a spike version of the Siamese network, was proposed based on the SiamFC [41-42]. The SiamSNN realized energy-efficient object tracking on the mainstream benchmarks including OTB2013 [44], OTB2015 (OTB-100) [45], VOT2016 [46], VOT2018 [47], and GOT-10k [48]. However, the tracking performance is still limited. It is urgent and highly desirable to explore high-performance SNN-based object trackers.

Here, we propose a Spiking SiamFC++ tracker based on IF neurons. Our contribution is three-fold. **First**, we design a novel framework of Spiking SiamFC++ tracker, and exploit the surrogate gradient to perform supervised training, which provides a novel path for SNN application in the field of target tracking. **Second**, the AlexNet network is extended in time domain, and the spike train is used for information transmission and calculation, so that the computation complexity of one forward propagation is lower and the time required is less in the inference process. **Third**, compared with the existed SNN-based target tracker, e.g., the SiamSNN, the precision of the proposed Spiking SiamFC++ is much higher, and the precision (succession) reaches 85.24% (64.37%). To our knowledge, it represents the best performance achieved by the SNN-based target tracker.

The rest of the paper is organized as follows. The network architecture and operation principle of the Spiking SiamFC++ is presented in section II. The spiking IF neuron model and surrogate gradient training algorithm are also presented. In Section III, the dataset and evaluation metric are introduced. In addition, the training and inference processes are presented in detail. The tracking performances for mainstream benchmarks are examined, and are compared to other existing popular object trackers. At last, conclusions are presented in section IV.

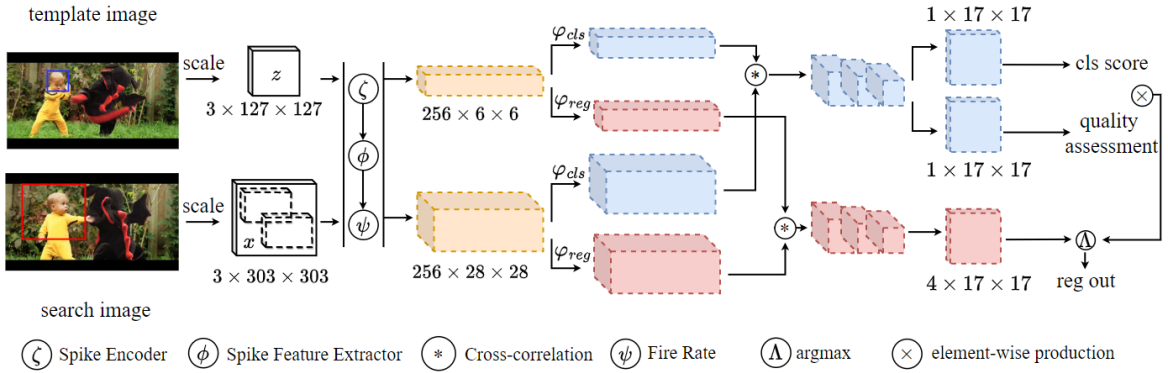

Fig.1 The architecture of the object tracker based on the proposed Spiking SiamFC++.

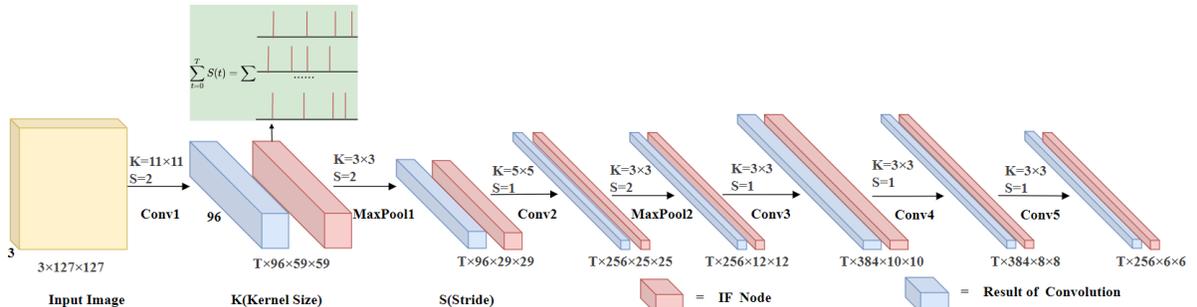

Fig.2. The architecture of Spiking AlexNet. K is the convolution kernel size, and S is the step size.

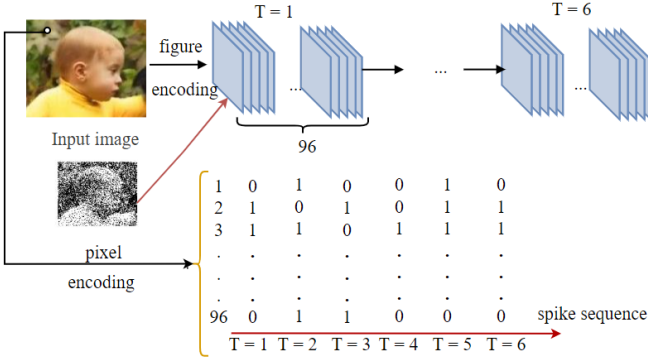

Fig.3. The schematic diagram of the spike encoding.

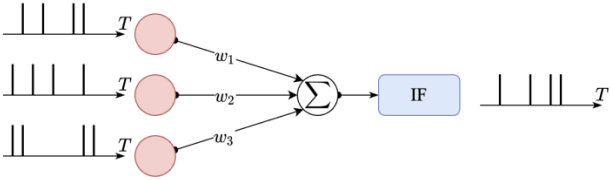

Fig.4. Time domain extension of SNN. IF represents the IF neuron.

## II. ARCHITECTURE AND MODEL OF SPIKING SIAMFC++

### A. Architecture of the Spiking SiamFC++

Here, Spiking SiamFC++ uses twin networks to solve the object tracking problem, which is similar to the SiamFC++. The architecture of the object tracker based on the proposed Spiking SiamFC++ tracker is presented in Fig.1. It can be decomposed into two sub-problems: the classification of the object and the accurate estimation of the object position. The classification (cls) and regression (reg) branches are considered in head.

During the network training, the template image and the search image are both input into the network simultaneously. As shown in Fig.1, the target marked position in the template frame is scaled to an RBG image with a size of $3 \times 127 \times 127$ (z), while the search area of the current search image is scaled to an RGB image with a size of $3 \times 303 \times 303$ (x). Then, both images are processed by spike encoder and spike feature extractor. Here, the spike encoder is used to encode an RGB image into a spike train of length T. Here, the observation time of the network is also represented by T. The common spike feature extractor is a time-domain extended version of AlexNet, denoted as Spiking AlexNet [49]. The architecture of the Spiking AlexNet is shown in Fig.2. It consists of multiple convolution layers, pooling layers, and spiking activation layers. The IF spiking neuron model is used to mimic the ReLU function and provide nonlinearity to the network. Essentially, the extension of SNN in time domain is to expand the original floating-point tensors into discrete spike tensors. In the original data stream, features are usually represented by 4-dimensional tensors. A representative example of the spike encoding of a single pixel is further presented in Fig.3 for T=6. After extending to time domain, there will be one more dimension T to represent spikes, as shown in Fig.3.

Spikes are then used to transmit information and participate in calculation. The remaining layers, such as convolution layer, pooling layer and fully connected layer, discrete binary sequences are used for calculation according to the original calculation method. The parameters of the Spiking AlexNet are presented in Table 1. For the template and search area, after the spike feature extractor, the size is $T \times 256 \times 6 \times 6$ and $T \times 256 \times 28 \times 28$, respectively. At the last layer, to remove the time domain, the firing rate of spiking neurons within the observation time T is calculated according to rate coding. The firing rate can be calculated as follows,

$$\psi = \frac{N}{T} \tag{1}$$

where $N$ denotes the spikes number and $T$ represents the observation time steps. If the neuron generates a spike at each time step, the maximum firing rate reaches 1.

**Table 1.** The network parameters for the Spiking AlexNet. B represents the Batch size of the network.

| Layer | Param | z'shape | x'shape |
|---|---|---|---|
| Conv2d | 34,944 | T, B, 96, 59, 59 | T, B, 96, 147, 147 |
| IFNode | 0 | T, B, 96, 59, 59 | T, B, 96, 147, 147 |
| MaxPool2d | 0 | T, B, 96, 29, 29 | T, B, 96, 73, 73 |
| Conv2d | 307,456 | T, B, 96, 29, 29 | T, B, 96, 73, 73 |
| IFNode | 0 | T, B, 256, 25, 25 | T, B, 256, 69, 69 |
| MaxPool2d | 0 | T, B, 256, 12, 12 | T, B, 256, 34, 34 |
| Conv2d | 885,120 | T, B, 256, 10, 10 | T, B, 384, 32, 32 |
| IFNode | 0 | T, B, 256, 10, 10 | T, B, 384, 32, 32 |
| Conv2d | 663,936 | T, B, 384, 8, 8 | T, B, 384, 30, 30 |
| IFNode | 0 | T, B, 384, 8, 8 | T, B, 384, 30, 30 |
| Conv2d | 442,624 | T, B, 256, 6, 6 | T, B, 256, 28, 28 |

After the common spike feature extraction, different convolution operations are further introduced for the cls branch and the reg branch. For both branches, three $3 \times 3$ convolution layers are used to further extract features. Then the cross-correlation coefficient between the template and search images is calculated for both the cls and the reg branches. The cross-correlation between the z and x can be expressed as [39],

$$f_i(z, x) = \varphi_i(\phi(\zeta(z))) * \varphi_i(\phi(\zeta(x))), i \in \{cls, reg\} \tag{2}$$

where $i$ denotes cls or reg branch, $\phi$ represents the spike feature extraction，$\zeta$ denotes spike encoder, $*$ is the convolution operation for two branches. The sizes of $\varphi_{cls}$ and $\varphi_{reg}$ are the same.

After the calculation of cross-correlation, we will calculate two parts for the cls branch including the cls score and quality assessment. The product of the cls score and quality assessment is regarded as the score of the predicted box. The size of the feature map becomes $1 \times 17 \times 17$ and $4 \times 17 \times 17$ for cls and reg branches, respectively.

For the cls branch, for, when mapping a position (x, y) from the extracted feature map back to the original image, if the mapped position $\left( \left\lfloor \frac{s}{2} \right\rfloor + xs, \left\lfloor \frac{s}{2} \right\rfloor + ys \right)$ is exactly covered by the ground truth (GT) bounding box, then it is regarded as a positive sample. Otherwise, it is regarded as a negative sample. Here, s=8 represents the stride step of the backbone network. For the

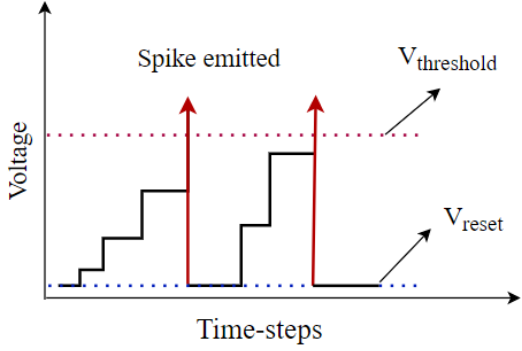

Fig.5. The schematic diagram of integration and fire process of an IF neuron.

reg branch, the distance between the mapped position $(\lfloor \frac{s}{2} \rfloor + xs, \lfloor \frac{s}{2} \rfloor + ys)$ and the GT bounding box can be represented with a 4-dimensional vector $d^* = (l, t, r, b)$, where

$l = (\lfloor \frac{s}{2} \rfloor + xs) - x_0$ , $t = (\lfloor \frac{s}{2} \rfloor + ys) - y_0$ , $r = x_1 - (\lfloor \frac{s}{2} \rfloor + xs)$ ,

$b = y_1 - (\lfloor \frac{s}{2} \rfloor + ys)$ , $(x_0, y_0)$ and $(x_1, y_1)$ represents the left-top

and right-bottom corners coordinates of the GT bounding box corresponding to the position (x, y). With this 4-dimensional vector $d*$, we can further calculate the prior spatial score (PSS) as follows [39],

$$PSS = \sqrt{\frac{\min(l, r)}{\max(l, r)} \times \frac{\min(t, b)}{\max(t, b)}} \quad (3)$$

Similar to the SiamFC++, when performing inference, the product of the cls score and the PSS represents the final score of the cls branch.

In addition, we also consider the IoU score as another quality assessment. The IoU can be calculated as follows [39],

$$IoU = \frac{Intersection(B, B^*)}{Union(B, B^*)} \quad (4)$$

where $B^*$ represents the GT bounding box, B denotes the predicted bounding box.

In the Spiking SiamFC++ tracker, we adopt the IF model as a spiking neuron. The detailed spiking neuron model can be referred to Refs. [41-42]. The schematic diagram of the integration and fire process of an IF neuron is illustrated in Fig.5.

### B. Surrogate gradient training algorithm

During training, the loss function of the Spiking SiamFC++ can be calculated as follows [39],

$$L = \frac{1}{N_{pos}} \sum_{x,y} L_{cls}(p_{x,y}, c_{x,y}^*) + \frac{1}{N_{pos}} \sum_{x,y} c_{x,y}^* L_{quality}(q_{x,y}, q_{x,y}^*)$$
$$+ \frac{1}{N_{pos}} \sum_{x,y} c_{x,y}^* L_{reg}(t_{x,y}, d_{x,y}^*) \quad (8)$$

where $N_{pos}$ represents the number of positive samples. $L_{cls}$ represents focal loss as classification result, $L_{quality}$ means the binary cross entropy loss for quality assessment, $L_{reg}$ is IoU loss for bounding box result. When (x, y) is considered as a positive (negative) sample, $c_{x,y}^* = 1$ ($c_{x,y}^* = 0$).

**Table 2.** The parameters used for training

| er | Description | Value |
|---|---|---|
| | Observation time window | 6 ms |
| | Simulation time step | 1 ms |
| | Membrane voltage threshold | 1 mV |
| | Optimizer | SGD |
| | Learning rate | 0.01 |
| | Batch size | 16 |

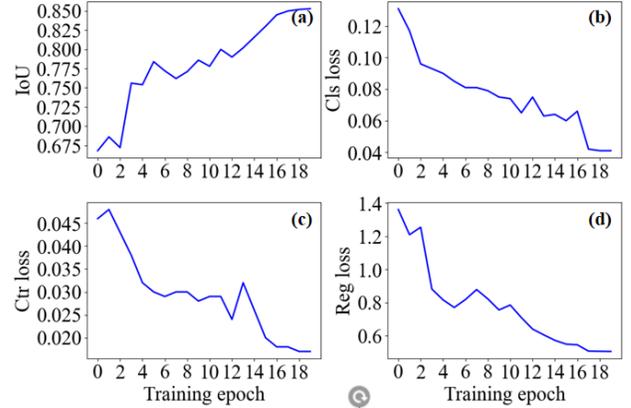

Fig.6. (a) IoU, (b) Cls loss, (c) Ctr loss, (d) Reg loss as a function of training epoch.

Note, to overcome the non-differentiable nature of spikes, we introduce surrogate gradient function to realize back propagation. More specifically, the step function is used in the forward propagation, while the surrogate functions are used to calculate the gradient in the back propagation. In our experiments, we adopted softsign as the surrogate gradient function as follows [28],

$$g(x) = \frac{1}{2} \left( \frac{\alpha x}{1 + |\alpha x|} + 1 \right) \quad (9)$$

## III. EXPERIMENTS AND RESULTS

To begin with, we present the dataset, then the training phase and the training algorithm of the proposed Spiking SiamFC++ is described. After training, the tracking results are analyzed by extensive experiments. The tracking performance for the proposed Spiking SiamFC++ is compared to that of the existing object tracking models.

### A. Dateset for the object tracker

Here, we adopt GOT-10K dataset as the training dataset in the training process of the Spiking SiamFC++, and evaluate our methods on OTB-2015(OTB-100) dataset. The GOT-10K dataset can be divided into training set, validation set and test set, containing 9935, 180 and 180 video sequences respectively. The average sequence length is about 150 frames. The dataset contains 563 kinds of objects in total, and the training set and validation set provide the label information of each frame. The OTB dataset is similar to the VOT dataset, except that it contains a quarter of the grayscale images.

## B. Training phase and the training algorithm

Experiments were carried out using the following hardware parameters. We adopted two CPUs (Intel(R) Xeon(R)E5-2620 V4, the operating frequency is 2.10GHz, the memory capacity is 64G, and the frequency is 2400MHz). In addition, two GPUs were used (Nvidia RTX 2070Super, the video memory is 8G×2). The software environment of this experiment was as follows, Python language (interpreter version 3.9), deep learning framework PyTorch (version 1.8.0) [50], and a SNN framework named SpikingJelly [51].

As mentioned above, surrogate gradient method is employed to train the network model. The surrogate gradient learning algorithm for the proposed Spiking SiamFC++ is shown in *Algorithm* 1. In addition, the parameters used during the training process are presented in Table 2. During the training phase, the evolutions of IoU, Cls loss, Ctr loss and the Reg loss are shown in Fig.6.

---

**Algorithm 1: Directly training Spiking SiamFC++ with surrogate gradient backpropagation**

---

**Input**: RGB image(3*127*127), T=6, max_epoch = 20
**Output**: Spiking Siamfc++ network
1: **for** epoch to max_epoch **do**
2:　**for** pixel in image **do**
3:　　**for** t to T **do**
4:　　　Spike Encoder → spike sequence (length = 96)
5:　　**end for**
6:　　pixel Spiking Encoder → spike sequence (T×96)
7:　**end for**
8:　image Spiking Encoder → feature map(T×96×59×59)
9:　Spike Feature Extractor (four layers convolution)→ feature map(T×256×6×6)
10:　Fire rate → feature map (256×6×6)
11:　classification and regression
12:　calculate loss
13:　Do back-propagation and weight update
14：**end for**

---

## C. Results and discussions

After the training process is converged, the trained model was firstly evaluated using the OTB100 dataset. Here, three object tracking models including the SiamFC, Siam FC++, and the proposed Spiking SiamFC++, are considered and compared in detail. To evaluate and compare quantitatively the performance achieved by different models, both the precision and success rate are calculated. Here, the precision can be calculated as [44],

$$precision = \frac{N_{distance < Local\ error\ threshold}}{N_{total}} \quad (10)$$

where $N_{total}$ represents the total number of frames in a video sequence, $N_{distance < Local\ error\ threshold}$ represents the total number of frames whose Euclidean distance is less than the defined local error threshold. Euclidean distance can be calculated as

$$distance = \sqrt{(x_0 - x_1)^2 + (y_0 - y_1)^2} \quad (11)$$

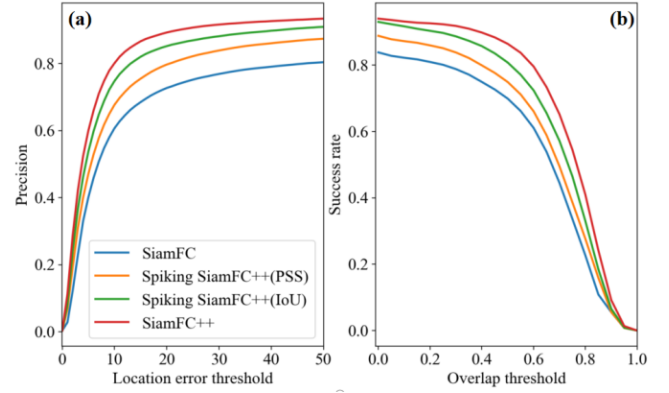

Fig.7. The plots of (a) precision and (b)success rate.

**Table 3.** Comparison of tracking performance of Spiking SiamFC++ to other trackers on OTB100.

| Model | Quality assessment | Success | Precision |
|---|---|---|---|
| SiamFC | - | 56.29 | 76.06 |
| SiamSNN | - | 44.32 | 52.78 |
| SiamFC++ | PSS | 65.60 | 88.40 |
| | IoU | 68.50 | 89.30 |
| **Spiking SiamFC++** | **PSS** | **62.49** | **82.09** |
| | **IoU** | **64.37** | **85.24** |

where $(x_0, y_0)$ represents the center point coordinates of the predicted frame and $(x_1, y_1)$ represents the center point coordinates of the GT bounding box. The success rate can be calculated as [44],

$$success\ rate = \frac{N_{IoU > Overlap\ threshold}}{N_{total}} \quad (12)$$

where $N_{IoU > Overlap\ threshold}$ represents the total number of frames whose IoU between the predicted box and the GT bounding box is greater than the overlap threshold.

The plots of precision and success rate for different object tracker models are presented in Fig.7. The results show that, the Spiking SiamFC++ shows better performance than the SiamFC. The performance of Spiking SiamFC++ is very close to that of the original SiamFC++. When the distance local error threshold is relatively small, as can be seen from Fig.7(a), the performance of the three models is almost the same. But with the increase of the distance local error threshold, the SiamFC++ performs better. That is, the precision of the Spiking SiamFC++ is slightly lower than that of the original SiamFC++ but higher than that of the SiamFC. For the success rate, as presented in Fig.7(b), a large overlap threshold leads to a low success rate. But when the conditions are relaxed, the success rate is different for a small overlap threshold, which also proves that the Spiking SiamFC++ performs slightly worse than the traditional SiamFC++, but performs better than the SiamFC.

The tracking performance achieved by the Spiking SiamFC++ is compared with other object tracking models for the OTB100 dataset, as presented in Table 3. Compared with the original SiamFC++, the Spiking SiamFC++ shows a slight decrease in performance, especially when IoU is used as quality assessment. The precision and success reached 85.24% and 64.37%, respectively, which was higher than 82.09% and 62.49% when PSS was used as quality assessment. While for the SNN-based object tracker named SiamSNN, the success and precision is 44.32% and 52.78, respectively. That is to say, compared with the SiamSNN, the proposed Spiking SiamFC++ shows much better tracking performance. Note, the inference speed is about 67 frames per second (FPS).

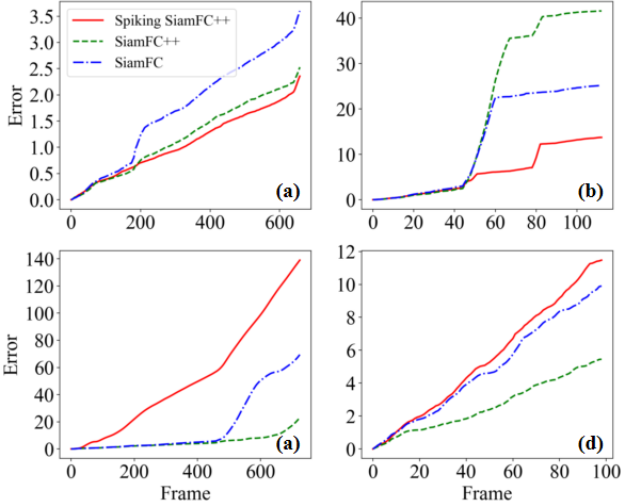

Fig.8. The error as a function of frame for (a) Car4, (b) DragonBaby, (c)Basketball, (d) Bird2.

Next, to illustrate the variation of error of each frame during tracking phase, we further calculate the error between the predicted bounding box $(x_{M,i}, y_{M,i})$ and the GT bounding box $(x_{gt,i}, y_{gt,i})$ as follows [52],

$$error = \frac{1}{N_f} \sum_{i=1}^{N_f} \sqrt{(x_{M,i} - x_{gt,i})^2 + (y_{M,i} - y_{gt,i})^2} \quad (13)$$

where $M$ denotes the tracking algorithm, $N_f$ is the total frame length, $i$ represents the $i$-th frame.

Four representative videos, Car4, DragonBaby, Basketball, and Bird4 sequences are considered here. Three object trackers are considered, namely, the SiamFC, SiamFC++, and the proposed Spiking SiamFC++. As shown in Fig. 8, in the first few frames, the three tracking algorithms exhibit similar error results, but with the increase of the number of frames, our algorithm outperforms SiamFC++ and SiamFC for the Car4 and DragonBaby sequences. For the Basketball sequence and Bird2 sequence, the performance of our algorithm is slightly lower than SiamFC++. We have also examined the variation of error extensively for other videos in the OTB100 dataset. According to statistical analysis of extensive experiments, we find that our algorithm performs better than the SiamFC++ for 32 sequences, but performs worse for 36 sequences. While for the remaining sequences, the performance is comparable for SiamFC++ and Spiking SiamFC++.

### D. Specific tracking performance

Object tracking is a very challenging task. The moving scene is extremely complex and may change at any time for moving objects. The main challenging factors in object tracking are as follows: object occlusion, deformation, background clutter, poor illumination and so on [34-35]. In the following, we consider these challenges to show the robustness of the proposed Spiking SiamFC++. 5 representative sequences from the OTB100 dataset are considered here. The corresponding visualized tracking results at different frames are presented in Fig.9.

**Drastic change of appearance**: Sequence 1 is selected from the DragonBaby in OTB100 dataset. During the tracking process, the target images move rapidly, leading to drastic changes in shape. As can be seen in Fig.9(a), the tracking box can be still closely following the target box, indicating that our method can solve the deformation problem to a certain extent.

**Object occlusion problem**: Sequence 2 is selected from Basketball in OTB100 dataset. Note, as presented in Fig.9(b), in frame 19 image, the target image is partially occluded, but in frame 37 image, the target image can still be tracked well. Sequence 3 is from FaceOcc1 in OTB100 dataset, as presented in Fig.9(c), although the target image is nearly half occluded, the predicted box can still cover the location of the target well, so our method can effectively alleviate the poor tracking effect caused by target occlusion.

**Background clutter problem**: Sequence 4 is selected from CarDark in OTB100 dataset. As shown in Fig.9 (d), there are similar objects around the target which cause interference to the tracking, but the tracker can still track the target, and the IoU is always above 0.7.

**Poor illumination problem**: Sequence 5 is selected from David in the OTB100 dataset. The results show that for poor ambient illumination condition, the IoU can remain above 0.9 as displayed in Fig.9(e), indicating that excellent tracking performance can be achieved even for poor ambient illumination.

### E. Tracking performance on other datasets

Without loss of generality, we also considered the tracking performance on some other benchmarks datasets, including OTB2015 [45], VOT2016 [46], VOT2018 [47] and UAV123 [53]. The tracking performance for the Spiking SiamFC++ is compared to the state-of-the-art object trackers. For the VOT dataset, three quantifiers, namely, expected average overlap (EAO), accuracy(A) and robustness (R) are considered [54]. For the UAV dataset, the area under curve (AUC) and distance precision (DP) are calculated [53]. Note that, a higher precision (success/accuracy) indicates better tracking performance. Robustness refers to the number of times that tracking fails under a single test sequence. A larger value of R means worse robustness. EAO reflects the relationship between sequence length and average accuracy, considering both accuracy and robustness. A higher EAO indicates better accuracy. The AUC and DP of UAV dataset are essentially the same as the Precision and Success of OTB dataset. The test results for the four benchmarks datasets with different object tracking models are displayed in Table 4.

**Results on OTB2015 Benchmark.** The OTB2015 dataset contains 100 video sequences, including 25% gray scale dataset. It can be seen that the precision (success) of our algorithm is 0.030 (0.036) lower than that of SiamFC++, but both quantifiers are higher than the SiamFC and SiamRPN.

**Results on VOT Benchmark.** We have conducted tests on VOT2016 and VOT2018 respectively. Similar to OTB2015 results, our results are better than SiamFC and SiamRPN. Each performance is similar to SiamFC++ and SiamRPN++. For VOT2018 dataset, the quantifier $A$ is the same as that for the SiamFC++.

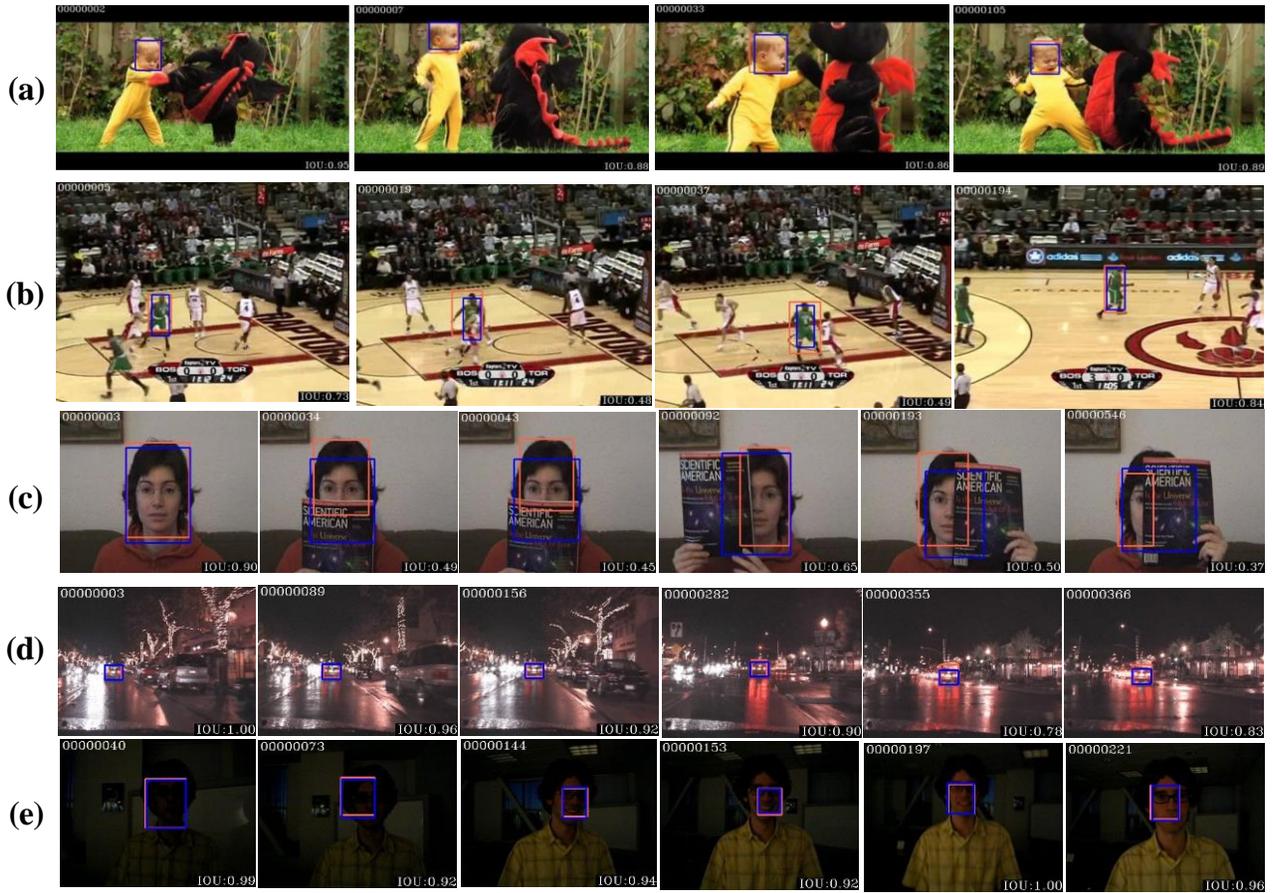

Fig.9. Tracking results of 5 representative sequences from OTB100 for (a) DragonBaby, (b) Basketball, (c) FaceOcc1, (d) CarDark, (e) David. The blue box is the predicted tracking box, and the red box is the GT bounding box. The serial number of the image in the video sequence is displayed in the upper left corner, and the IoU in the current frame is displayed in the lower right corner.

Table 4. Comparison of tracking performance of Spiking SiamFC++ to other trackers for different benchmarks.

| Trackers | | SiamFC (2016) | SiamRPN (2018) | SiamRPN++ (2019) | SiamFC++ (2020) | SiamSNN (2021) | Spiking SiamFC++ (ours) |
|---|---|---|---|---|---|---|---|
| OTB2015 | Precision | 0.761 | 0.851 | 0.914 | 0.884 | 0.528 | **0.854** |
| | Success | 0.563 | 0.637 | 0.696 | 0.682 | 0.443 | **0.644** |
| VOT2016 | A | 0.529 | 0.560 | 0.612 | 0.626 | 0.497 | **0.600** |
| | R | 0.460 | 0.260 | 0.266 | 0.144 | 0.630 | **0.359** |
| | EAO | 0.235 | 0.344 | 0.357 | 0.460 | 0.210 | **0.302** |
| VOT2018 | A | 0.503 | 0.490 | 0.600 | 0.556 | 0.460 | **0.556** |
| | R | 0.585 | 0.460 | 0.234 | 0.183 | 0.860 | **0.445** |
| | EAO | 0.188 | 0.244 | 0.414 | 0.400 | 0.176 | **0.255** |
| UAV123 | AUC | 0.504 | 0.527 | 0.578 | 0.623 | - | **0.578** |
| | DP | 0.702 | 0.748 | 0.769 | 0.781 | - | **0.744** |

**Results on UAV123 Benchmark.** The UAV123 dataset consists of 123 video sequences, and the average length is 915 frames. It is a dataset with specialized scenes, which are collected by unmanned aerial vehicle (UAV). Our algorithm achieves the same performance as SiamRPN++ in AUC, and is better than the SiamFC and SiamRPN.

In general, by comparing the results shown in Table 4, the proposed Spiking SiamFC++ performs well on all the considered datasets. Thus, the Spiking SiamFC++ is helpful and valuable for extending the potential applications of deep SNN in the field of object tracking.

## IV. CONCLUSION

In summary, we proposed a framework of SNN-based object tracker named Spiking SiamFC++ with the help of the surrogate gradient direct supervised training. The experiments show that, compared with the original SiamFC++, the precision loss is relatively small for the OTB100 benchmark dataset. On the other hand, compared with the SiamSNN, the precision is improved substantially, the precision and success reached 85.24% and 64.37%, respectively. To our knowledge, the presented Spiking SiamFC++ tracker performs better than the existed state-of-the-art SNN-based object trackers. Besides, high tracking performance can be obtained even for the videos with object occlusion, deformation, background clutter and poor illumination. The extensive experiments further show that the Spiking SiamFC++ also achieves high performance for other popular object tracking tasks, which verifies its tracking and generalization capability. We believe that this approach is of great significance for the future potential applications of SNN in target tracking and other computer vision tasks on power-efficient neuromorphic electronic and photonic SNN chips [4-11].

**Licun Yu** was born in Ningbo, China, in 1985. He is currently working toward the Ph.D. degree from Chang'an University, Xi'an, China. His research interests include object detection and object tracking.

**Yuechun Shi** was born in Nantong, China, in 1983. He received the Ph.D. degree from Nanjing University, Nanjing, China, in 2012.

In 2016-2022, he is an associate Professor with the College of Engineering and Applied Sciences, Nanjing University, Nanjing, China.

He is currently the researcher fellow with Yongjiang laboratory, Ningbo, China. His current research interests include semiconductor laser array, integrated photonics, and photonic neuromorphic computing.

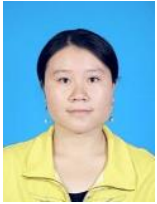

**Shuiying Xiang** was born in Ji'an, China, in 1986. She received the Ph.D. degree from Southwest Jiaotong University, Chengdu, China, in 2013.

She is currently a Professor with State Key Laboratory of Integrated Service Networks, Xidian University, Xi'an, China. Her research interests include neuromorphic photonic systems, spiking neural network, and semiconductor lasers dynamics.

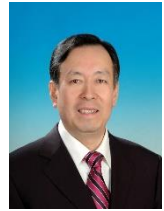

**Yue Hao** was born in the city of Chongqing, China, in 1958. He received the Ph.D. degree from Xi'an Jiao tong University, Xi'an, China, in 1991.

He is currently a Professor at State Key Discipline Laboratory of Wide Bandgap Semiconductor Technology, Xidian University, Xi'an, China. His research interests include semiconductor materials and devices.

**Tao Zhang** was born in Gansu Province, China, in 1999.He is currently working toward the master's degree with Xidian University, Xi'an, China. His research interests include spiking neural network.

**Shuqing Jiang** was born in Henan Province, China, in 1996.He is currently working toward the master's degree with Xidian University, Xi'an, China. His research interests include spiking neural network.

**Yanan Han** was born in Ningxia Hui Autonomous Region, China, in 1996. She is currently working toward the Ph.D. degree with Xidian University, Xi'an, China. Her research interests include neuromorphic photonic systems and spiking neural network,.

**Yahui Zhang** was born in Zhangjiakou, China, in 1993. She received the Ph.D. degree from Xidian University, Xi'an, China. Her research interests include vertical cavity surface emitting lasers, neuromorphic photonic systems, and spiking neural networks.

**Chenyang Du** was born in Shanxi Province, China, in 2000. He is currently working toward the master's degree with Xidian University, Xi'an, China. His research interests include spiking neural network.

**Xingxing Guo** was born in Ji'an, China, in 1993. She received the Ph.D. degree from Xidian University, Xi'an, China. Her research interests include neuromorphic photonic systems.